\documentclass[10pt,twocolumn,letterpaper]{article}

\usepackage[pagenumbers]{wacv} 

\definecolor{wacvblue}{rgb}{0.21,0.49,0.74}
\usepackage[pagebackref,breaklinks,colorlinks,allcolors=wacvblue]{hyperref}

\usepackage{graphicx}
\usepackage{booktabs}

\usepackage{multirow}
\newcommand{\Sim}{\emph{MostSimilarFrame}\xspace}
\newcommand{\Last}{\emph{LastCleanFrame}\xspace}
\newcommand{\Previous}{\emph{PreviousFrame}\xspace}
\newcommand{\Random}{\emph{RandomFrame}\xspace}
\newcommand{\irref}{\emph{IrRef}\xspace}
\newcommand{\noref}{\emph{NoRef}\xspace}
\newcommand{\reref}{\emph{Ref}\xspace}

\newcommand{\papertitle}{\emph{RefGlitch-Bench}\xspace}
\newcommand{\godot}{\emph{RefGlitch-Bench}\xspace}

\usepackage[most]{tcolorbox}

\newtcolorbox{rqbox}{
  colback=black!2,
  colframe=black!30,
  boxrule=0.4pt,
  arc=3pt,              
  left=4pt,right=4pt,top=3pt,bottom=3pt,
  boxsep=2pt
}

\def\wacvPaperID{1446} 
\def\confName{WACV}
\def\confYear{2027}

\title{\papertitle: A Benchmark for Reference-based Gameplay Glitch Detection with Vision-Language Models}

\author{Yakun Yu\\
University of Alberta\\
\and
Ashley Wiens\\
University of Alberta\\
\and
Adrián Barahona-Ríos\\
Sony Interactive Entertainment\\
\and
Benedict Wilkins\\
Sony Interactive Entertainment\\
\and
Saman Zadtootaghaj\\
Sony Interactive Entertainment\\
\and
Nabajeet Barman\\
Sony Interactive Entertainment\\
\and
Cor-Paul Bezemer\\
University of Alberta
}

\begin{document}
\maketitle
\begin{abstract}
Visual glitches in video games degrade player experience and perceived quality, yet manual quality assurance cannot keep pace with the growing test surface of modern game development.
Prior automation efforts, particularly those using vision-language models (VLMs), largely operate on isolated frames without sufficient context to judge whether a glitch is present.
We introduce \papertitle, a benchmark for reference-based video game glitch detection with VLMs. The key idea is to formulate glitch detection as an explicit within-video comparison problem: given a test frame, a reference frame provides a visual baseline that helps the model distinguish true glitches from benign visual variation. 
\papertitle includes a controlled synthetic dataset with five injected glitch types and manually annotated reference/test frame pairs, enabling an oracle-reference evaluation that isolates the potential benefit of reference guidance. We further establish four initial baselines for automatically selecting references from earlier frames in the same video, with \Last performing best and transferring across VLMs. Finally, we evaluate automatic reference guidance on real-world gameplay data, where it improves frame-level glitch detection beyond the controlled setting while revealing reference reliability and error propagation as key challenges.

\end{abstract}
    
\section{Introduction}
\label{sec:intro}
\begin{figure*}[tb]
  \centering
  \includegraphics[width=0.95\textwidth]{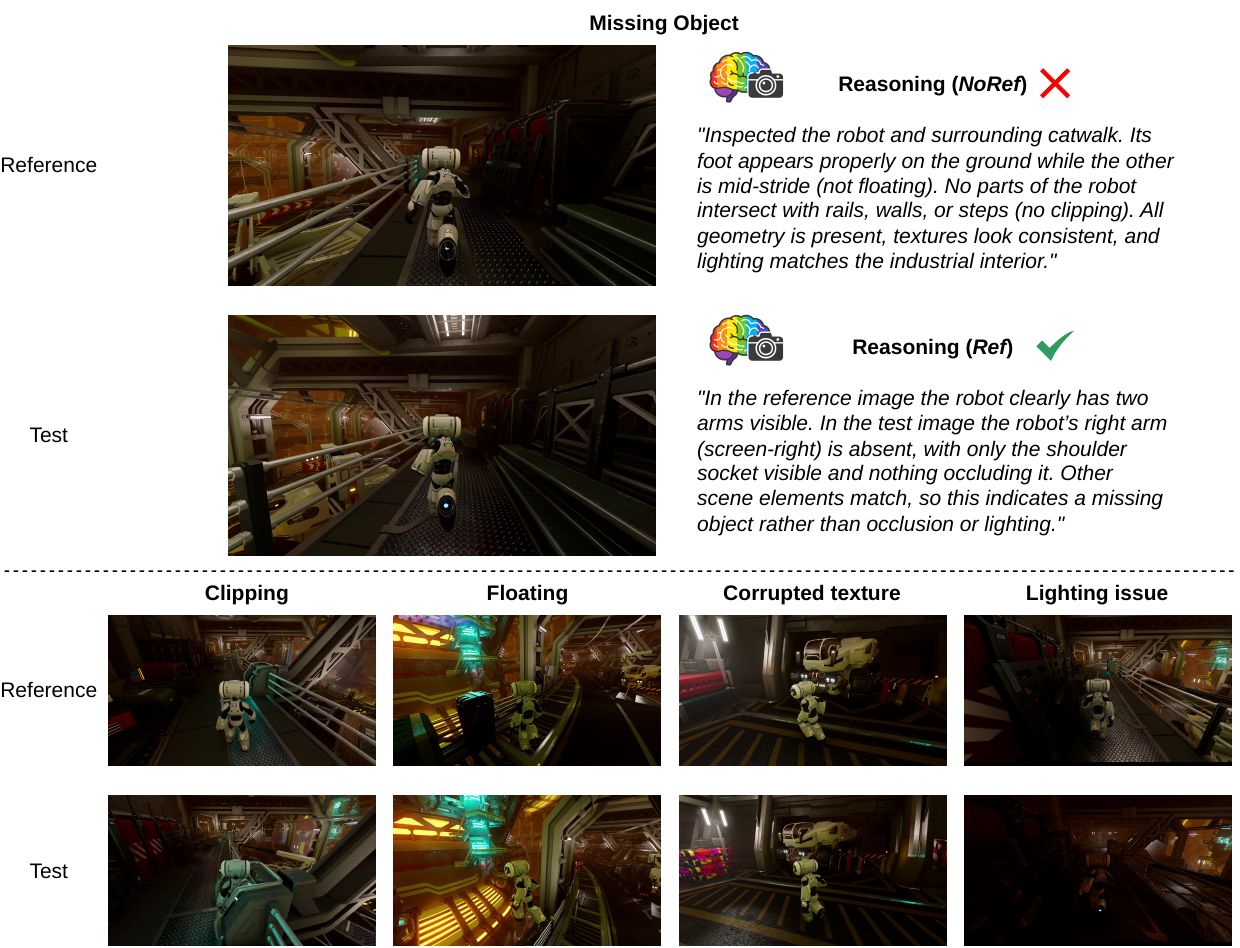}
  \caption{
    Reference/test pairs from the synthetic \godot dataset. Top: a \textbf{missing-object} (parts of a character/object are unexpectedly absent) example with GPT-5’s judgments with (\reref) and without (\noref) a reference. Bottom: paired examples for \textbf{clipping} (a character/object intersects or passes through a solid surface), \textbf{floating} (a character/object is not in contact with the surface it should rest on), \textbf{corrupted texture} (visibly broken or incorrect surface texturing), and \textbf{lighting issue} (abnormal scene illumination not explainable by a natural change).}
  \label{fig:data}
\end{figure*}

Video games are the most widely consumed form of interactive media, but their increasingly complex visuals and frequent content updates also introduce a steady stream of visual glitches. Glitches such as clipping, corrupted UI elements, and abnormal lighting undermine perceived quality and player trust, and can trigger negative reviews, refund requests, and costly customer support interactions~\cite{backus2025players, lin2019empirical, lu2025automated}. 
At the same time, manual quality assurance (QA) is becoming increasingly costly and time-consuming.
Modern games ship frequent builds across platforms, hardware configurations, and graphics settings, which makes exhaustive human review infeasible and expands the testing surface beyond what teams can reliably cover~\cite{taesiri2024searching, taesiri2022clip, chang2019reveal}.
Together, these pressures create a clear demand for accurate and scalable automated gameplay glitch detection.

Prior research on automated gameplay glitch detection, particularly those using VLMs, has largely operated in a single-image setting: given one screenshot, the model predicts whether it contains a visual artifact~\cite{ling2020using, taesiri2024glitchbench, taesiri2025videogamebunny, paduraru2026state}. 
This paradigm is inherently limited for accurate and contextual QA. First, a single image often lacks sufficient context to distinguish true glitches from benign variation (\eg, camera motion or stylistic effects). 
Second, many glitches are relative or temporal in nature, and are therefore difficult to diagnose from a single screenshot. For example, GPT-5 predicts the missing-object test frame in~\Cref{fig:data} as glitch-free when evaluated in isolation, because the missing robot arm is not obvious without a reference.

These limitations motivate a reference-guided view of gameplay glitch detection. Rather than asking a VLM to judge a test frame in isolation, we formulate glitch detection as an explicit within-video comparison problem: given a test frame, the model is also provided with a reference frame that serves as an expected visual baseline. In the example in~\Cref{fig:data}, providing a reference frame with an intact robot in a similar scene makes the discrepancy clear, allowing the VLM to correctly detect the missing right arm. This design aligns with how humans often perform QA: we identify visual failures by comparing the current scene against what the same object, lighting, geometry, or layout should look like under normal gameplay. Moreover, existing benchmarks do not provide a paired reference/test structure needed to study the impact of a reference image.

Therefore, we introduce \papertitle, a benchmark for reference-based video game glitch detection with VLMs. \papertitle is a controlled synthetic dataset built from an open-source game environment with five injected glitch types. Unlike existing multi-game benchmarks that mix diverse visual styles and provide limited per-category samples, \papertitle offers consistent rendering conditions and balanced coverage across glitch categories. Crucially, it includes manually labeled reference/test frame pairs, enabling controlled evaluation of reference guidance and performing fine-grained category-level analysis of which glitches benefit most from reference-based comparison. 

The manually constructed pairs define an oracle-reference condition. Each test frame is paired with a manually selected, visually related, and verified glitch-free reference. This condition deliberately removes reference selection error so that we can isolate the potential value of reference guidance itself. It should therefore be interpreted as an empirical upper-bound condition under our fixed single-reference evaluation protocol, rather than as a realistic deployment assumption or a solution to reference selection. 
We then define an open automatic-reference-selection setting in which a method selects a reference from earlier frames in the same video to aid the prediction of a test frame. We provide four initial baselines: \Last, \Previous, \Random, and \Sim. Among these simple policies, \Last, which selects the most recent preceding frame predicted to be glitch-free, performs best. Nevertheless, it remains below the oracle-reference condition and depends on potentially incorrect predictions about earlier frames. 

We further evaluate automatic reference guidance on frames from real-world gameplay videos. This experiment examines whether the frame-level benefits observed in the controlled benchmark transfer to more diverse visual conditions and, crucially, how the correctness of a selected reference affects its corresponding test-frame prediction.
In summary, we make the following major contributions: 
\begin{itemize} 

\item We introduce \papertitle, a controlled benchmark for reference-based frame-level gameplay glitch detection, comprising complete synthetic videos, and balanced reference/test pairs. It can support future research on both improved frame-level reference selection and the separate problem of aggregating frame-level signals into reliable video-level decisions. 

\item We define an oracle-reference condition that uses manually scene-matched and verified glitch-free references to remove reference-selection error and isolate the potential effect of reference guidance across various VLMs.

\item We establish an open automatic-reference-selection setting, provide several initial baselines, and identify \Last as the strongest among the simple policies tested, while retaining the oracle condition as an empirical upper bound.

\item We evaluate automatic reference guidance on frame-level examples from real-world gameplay videos and quantify how incorrect references propagate to subsequent frame predictions, identifying reliable reference retrieval and verification as central open problems.

\end{itemize}

\section{Related Work}

\subsection{Gameplay Glitch Detection Benchmarks} 
Existing resources for gameplay glitch detection can be broadly grouped into image-level and video-level benchmarks. Image-level benchmarks evaluate whether models can identify visual glitches from individual screenshots. For example, GlitchBench~\cite{taesiri2024glitchbench} introduced a benchmark of glitch-free and glitchy screenshots from many games to evaluate whether large multimodal models can detect image-level anomalies. VideoGameBunny~\cite{taesiri2025videogamebunny} further expands game-image understanding by releasing a large-scale collection of gameplay images and image-instruction pairs across hundreds of titles. 
These datasets have played an important role in establishing gameplay glitch detection as a multimodal reasoning problem, but they primarily evaluate glitches from isolated frames and therefore do not directly test whether additional visual context improves detection. 

Video-level benchmarks extend this setting by evaluating glitches over gameplay clips, but they remain comparatively limited. 
Several approaches~\cite{guglielmi2023using, truelove2023finding, article} first segment gameplay videos using auxiliary metadata such as streamer captions, and then classify segments as glitchy or non-glitchy. However, such metadata is often unavailable during game development. 
More recent benchmarks study gameplay videos directly. PhysGame~\cite{cao2024physgame} focuses on physical commonsense violations in gameplay videos. 
VideoGameQA-Bench~\cite{taesiri2025videogameqabench} includes video-level glitch detection as part of a broader video game QA benchmark. 
TempGlitch~\cite{yu2026tempglitch} introduces a controlled gameplay video benchmark specifically for temporal glitch detection.
Together, these datasets advance video-based gameplay understanding, but they do not explicitly provide paired reference/test frames for studying whether a visual baseline helps glitch detection. 

\papertitle is complementary to these benchmarks. Rather than evaluating models only on isolated screenshots or unpaired frame sequences, \papertitle introduces a reference-based setting in which each test frame is paired with a reference frame from the same video.
This design enables controlled analysis of whether visually related context helps VLMs distinguish true glitches from benign scene variation. Moreover, existing benchmarks often span many games and mix diverse visual styles. Their samples within any single game environment or glitch category are also limited. In contrast, \papertitle provides balanced coverage across five glitch categories within a consistent game environment, enabling fine-grained category-level analysis of glitches.

\subsection{Vision-Language Models for Game QA}

VLMs have rapidly evolved from contrastive image--text pretraining~\cite{pmlr-v139-radford21a, jia2021scaling, zhai2022lit, zhai2023sigmoid, li2021align} into instruction-following multimodal assistants~\cite{dai2023instructblip, chen2024internvl, zhang2026instruction, lee2024instruction}, with downstream adaptation typically achieved through prompt engineering~\cite{yu2024peprec, besta2024graph, wei2022chain}, instruction tuning~\cite{zhang2026instruction, dai2023instructblip, lee2024instruction}, and parameter-efficient fine-tuning~\cite{houlsby2019parameter, hu2022lora}.
Closed-source systems such as GPT-family models~\cite{singh2025openai} and Gemini~\cite{team2023gemini} demonstrate strong multimodal reasoning capabilities, while open-weight families such as Qwen-VL~\cite{bai2023qwen} and Ministral~\cite{liu2026ministral} provide increasingly competitive alternatives with lower deployment cost. These capabilities have motivated growing interest in applying VLMs to game QA, including glitch detection, and physical commonsense evaluation in gameplay videos. Despite this progress, current evidence suggests that VLM-based gameplay glitch detection remains far from solved. GlitchBench~\cite{taesiri2024glitchbench} and VideoGameBunny~\cite{taesiri2025videogamebunny} show that prompting-based methods can struggle with visual glitches across diverse games and scenes. VideoGameQA-Bench~\cite{taesiri2025videogameqabench} reports substantial headroom for video glitch detection, and PhysGame~\cite{cao2024physgame} shows that even strong VLMs have difficulty identifying physical commonsense violations in gameplay videos. These findings suggest that simply applying general-purpose VLMs to gameplay data is insufficient for robust QA automation. 

These limitations point to an evaluation gap, rather than solely a modeling gap. \papertitle addresses this gap while retaining a VLM-based evaluation setting by introducing a new dataset with manually selected reference frames for reference-based evaluation. It further establishes four initial baselines for automatic reference selection, providing a foundation for future work on more effective reference-selection strategies.

\section{\papertitle Benchmark}

Existing benchmarks~\cite{taesiri2025videogamebunny, taesiri2024glitchbench, cao2024physgame, taesiri2025videogameqabench} span many games, rendering styles, and camera behaviors, and they rarely provide explicit reference frames. As a result, it is difficult to isolate the effect of reference guidance. Moreover, analysis across glitch categories is often underpowered due to limited samples within any single game. To address these gaps, we introduce \papertitle, a controlled benchmark specifically designed for reference-based glitch detection, featuring (i) synthetic gameplay videos with manually annotated reference/test frame pairs, (ii) balanced coverage of five glitch types within a single, consistent game environment, and (iii) four initial baselines for automatic reference selection and reference-based detection.

Concretely, we used the open-source Godot engine and the Third Person Shooter (TPS) demo asset
as a base environment. We modified the source code to implement each glitch with fast in-engine toggling (See the supplementary material for implementation details), enabling recordings under the same camera and rendering conditions. We generated five glitch types: 

\begin{enumerate}
    \item \textbf{Missing object}: parts of the character or object are unexpectedly missing.
    \item \textbf{Clipping}: a character or object visibly intersects or passes through another solid surface, or geometry overlaps in a way that could not occur with correct depth/occlusion.
    \item \textbf{Floating}: a character or object is visibly not in contact with the ground or surface it should be resting on, defying expected physics or gravity.
    \item \textbf{Corrupted texture}: a surface shows clearly broken or incorrect texturing. 
    \item \textbf{Lighting issue}: the scene lighting is clearly wrong and not explainable by natural scene change.
\end{enumerate}

We recorded gameplay videos of 5 to 18 seconds with injected glitches. Each video clip starts with normal behavior. About one-third into the video, the glitch is triggered and remains active for several seconds. A few seconds before the end, the game returns to normal behavior. We recorded 100 videos for each glitch type. For each video, we sampled frames at 5 FPS to cover both glitch-free and glitchy intervals. 
Two annotators manually constructed the pairs by first identifying a test frame in which the target glitch was clearly visible or a clearly glitch-free frame for negative examples, and then selecting a temporally nearby, verified glitch-free frame from the same video with a similar scene configuration and a clear visual contrast to the test frame. 
Following this procedure, we collected 500 pairs with a glitch and 500 without a glitch, yielding 1,000 paired samples in total.
\Cref{fig:data} shows example pairs for each glitch type.

\section{Automatic Reference Selection}
\begin{figure}[tb]
  \centering
  \includegraphics[width=\columnwidth]{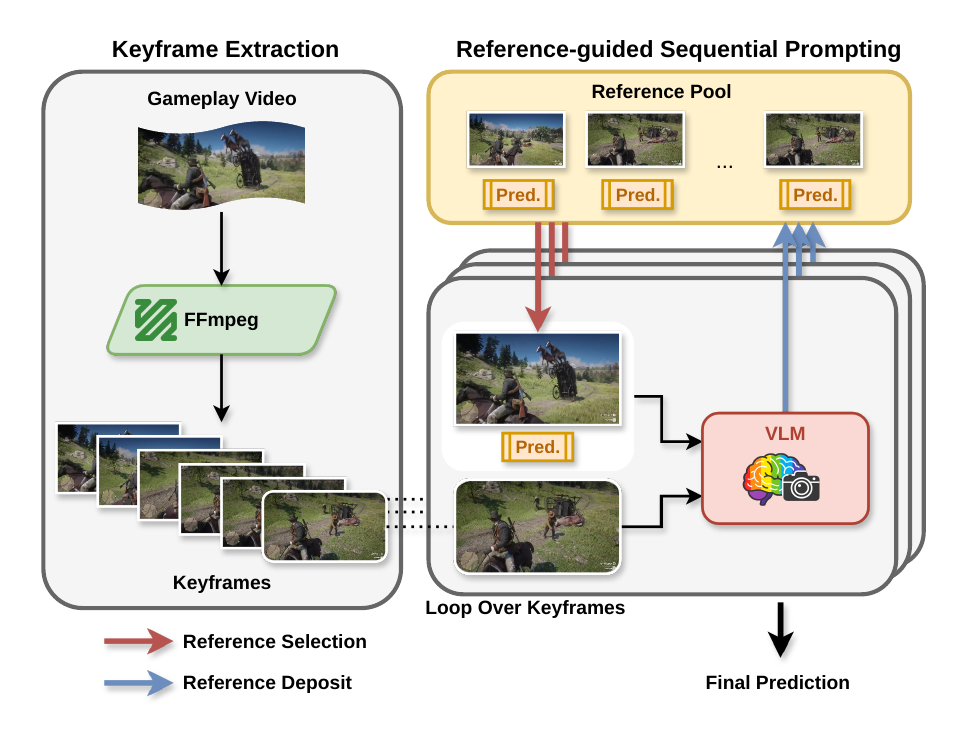}
  \caption{Overview of our automatic reference selection method that consists of two stages. (1) Keyframe extraction: we extract a compact set of representative frames. (2) Reference-guided sequential prompting: we process frames in temporal order and prompt a VLM to predict whether the current test frame is glitchy, conditioned on an additional reference frame selected from a reference pool of earlier frames. Each stored reference is associated with its predicted label (``Pred.''); red arrows denote reference selection and blue arrows denote depositing newly processed frames back into the pool.}
  \label{fig:framework}
\end{figure}

Manually selected reference frames establish an informative upper bound on how much a visual baseline can help gameplay glitch detection, but selecting references manually is not practical at scale. Therefore, we establish a baseline approach for automatic reference selection that allows VLMs to be evaluated on gameplay videos when manual reference labels are not available. As shown in~\Cref{fig:framework}, the approach consists of two stages. First, we extract representative keyframes from each gameplay video. Second, we process keyframes sequentially and obtain frame-level glitch predictions via reference-guided prompting. 

\paragraph{Keyframe Extraction.}
Gameplay videos often contain substantial frame-to-frame redundancy, making direct dense-frame analysis inefficient and computationally expensive. To make detection tractable, we reduce each video into a compact set of $T$ keyframes.
Specifically, we extract I-frames (intra-coded frames) from the video using \textit{FFmpeg}~\cite{ffmpeg}. I-frames are self-contained and can be decoded independently, yielding a sparse sequence of representative frames determined by the video’s encoding settings.

\paragraph{Reference-guided Sequential Prompting.}
Each keyframe is evaluated via reference-guided prompting, where the VLM judges the current frame using contextual evidence from an earlier frame in the same video.
Specifically, for each extracted keyframe, we select a reference frame from a \emph{reference pool} that stores earlier frames along with their predicted labels (``Pred.'') as illustrated in \Cref{fig:framework}. For the first frame $I_1$, the pool is empty, so we use an adapted version of the prompt (see the supplementary material) that takes a single frame and asks the VLM to determine whether $I_1$ contains any glitch evidence based solely on observable visual cues. Once a prediction is produced, $I_1$ and its predicted label are deposited into the pool.
Starting from keyframe $t\ge 2$, we perform reference-guided prompting. For each test frame $I_t$, we select a reference frame $R_t$ from the pool according to a reference selection policy (described in~\Cref{sec:RQ2}). We then prompt the VLM with the pair $(R_t, I_t)$.  
The VLM outputs a frame-level prediction $\hat{y}_t$ (glitchy vs.\ glitch-free) together with a brief rationale describing how the reference supports the decision.

Since a glitchy reference would otherwise mislead the model, the prompt is conditioned on $R_t$'s predicted status. If $R_t$ is predicted glitch-free, the VLM is instructed to treat it as a normal baseline and flag visual discrepancies in $I_t$ as evidence of a glitch. If $R_t$ is predicted glitchy, the prompt warns the model accordingly and asks it to judge whether $I_t$  also contains evidence of a glitch.
After prediction, $I_t$ and its predicted label are deposited into the reference pool (blue arrows in \Cref{fig:framework}), so that subsequent frames can draw on a reference that is temporally close to them.
\section{Experimental Setup}

\paragraph{Real-world Dataset.} To evaluate whether reference guidance improves glitch detection under realistic QA conditions, we further evaluated it on two state-of-the-art real-world gameplay video benchmarks: PhysGame~\cite{cao2024physgame} and VideoGameQA-Bench~\cite{taesiri2025videogameqabench}.
PhysGame targets physical commonsense violations in gameplay footage and contains 880 videos with glitches from diverse games, spanning four domains (mechanics, kinematics, optics, and material properties) across 12 physical commonsense categories. 

Because PhysGame contains only glitchy clips, we complemented it with 500 glitch-free videos from VideoGameQA-Bench. 
To match the size of our benchmark, we randomly sampled 87 videos and annotated 1,000 frames, comprising 500 glitchy and 500 glitch-free frames.
Annotators assigned a binary label to each frame based on visible evidence: a frame was labeled glitchy only when a glitch was visually present, and otherwise labeled glitch-free, even when it came from a video labeled positive by PhysGame.


\paragraph{Evaluation Metrics.}
We evaluate glitch detection as a binary classification task, where each frame is labeled as glitchy or glitch-free. We report the performance using standard classification metrics: accuracy, F1, precision and recall. We primarily use accuracy and F1 as headline metrics for model comparison. We report precision and recall not as standalone optimization targets, but as diagnostic metrics that help interpret performance differences.

\paragraph{VLM Backbones and Compute.}
We evaluate five VLMs, including two proprietary models and three open-weight models. The proprietary models are GPT-5~\cite{singh2025openai} and Gemini 3 Flash Preview~\cite{team2023gemini}. The open-weight models are Qwen3-VL-8B~\cite{bai2023qwen}, Ministral-3-8B~\cite{liu2026ministral}, and Gemma-3-12B~\cite{gemmateam2025gemma3technicalreport}. For proprietary models, we use the official APIs. For open-weight VLMs, we use the public Unsloth AI framework. All experiments with open-weight models are run on 4 NVIDIA L40S GPUs.
\section{Results}
\label{sec:results}
\subsection{RQ1: Does a manually selected reference frame improve frame-level glitch detection?} \label{sec:RQ1}

\begin{table}[t]
\centering
\scriptsize
\setlength{\tabcolsep}{3pt}
\renewcommand{\arraystretch}{1.03}
\begin{tabular}{llcccc}
\toprule
\textbf{Model} & \textbf{Setting} & \textbf{Accuracy} & \textbf{F1} & \textbf{Precision} & \textbf{Recall} \\
\midrule
\multirow{4}{*}{GPT-5}
& \reref   & 0.77 & 0.74 & 0.87 & 0.64 \\
& \Last & 0.70 & 0.62 & 0.85 & 0.48 \\
& \noref & 0.65 & 0.49 & 0.88 & 0.34 \\
& \irref & 0.66 & 0.51 & 0.88 & 0.36 \\
\cmidrule(lr){2-6}
\multirow{4}{*}{Gemini 3 Flash Preview}
& \reref   & 0.63 & 0.73 & 0.58 & 0.98 \\
& \Last & 0.59 & 0.68 & 0.56 & 0.86 \\
& \noref & 0.67 & 0.65 & 0.70 & 0.61 \\
& \irref & 0.68 & 0.65 & 0.72 & 0.60 \\
\midrule
\multirow{4}{*}{Qwen3-VL-8B}
& \reref   & 0.73 & 0.74 & 0.72 & 0.76 \\
& \Last & 0.70 & 0.69 & 0.70 & 0.69 \\
& \noref & 0.62 & 0.42 & 0.88 & 0.28 \\
& \irref & 0.54 & 0.19 & 0.83 & 0.11 \\
\cmidrule(lr){2-6}
\multirow{4}{*}{Ministral-3-8B}
& \reref   & 0.68 & 0.58 & 0.83 & 0.45 \\
& \Last & 0.63 & 0.45 & 0.86 & 0.30 \\
& \noref & 0.57 & 0.35 & 0.73 & 0.23 \\
& \irref & 0.50 & 0.03 & 0.44 & 0.02 \\
\cmidrule(lr){2-6}
\multirow{4}{*}{Gemma-3-12B}
& \reref   & 0.63 & 0.63 & 0.63 & 0.63 \\
& \Last & 0.61 & 0.60 & 0.62 & 0.58 \\
& \noref & 0.58 & 0.48 & 0.64 & 0.38 \\
& \irref & 0.51 & 0.14 & 0.60 & 0.08 \\
\bottomrule
\end{tabular}
\caption{Overall performance under different reference settings across different VLMs. The \Last setting is discussed in RQ2.}
\label{tab:RQ1}
\end{table}

\begin{figure*}[tb]
  \centering
  \includegraphics[width=\textwidth]{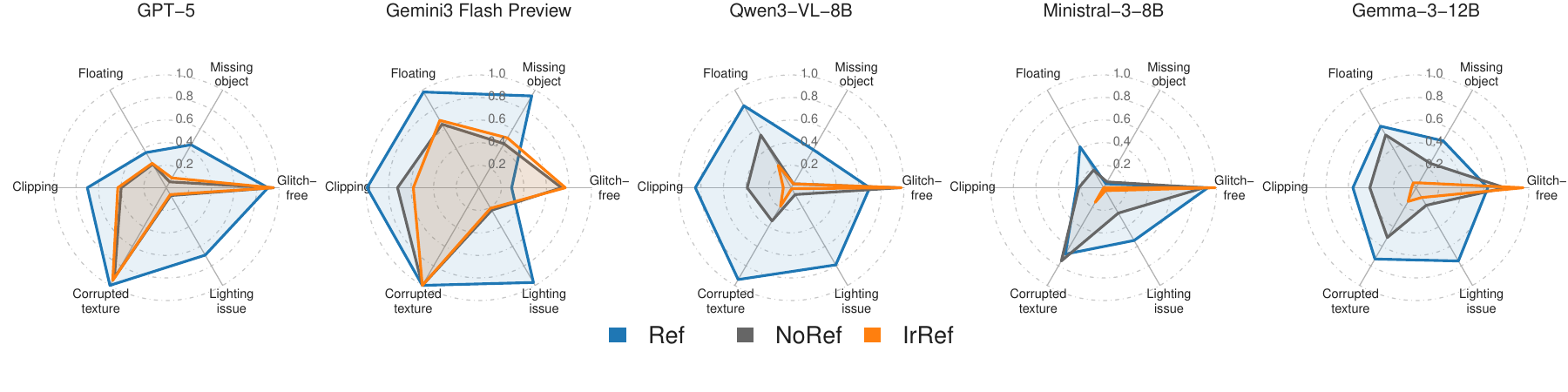}
  \caption{Per-category behavior of the VLMs visualized as radar plots over six classes (Glitch-free, Missing object, Clipping, Floating, Corrupted texture, Lighting issue), with lines color-coded by settings. Please zoom-in for a better view.}
  \label{fig:overall_performance}
\end{figure*}

\paragraph{Motivation.}
Many gameplay glitches are subtle and easily confused with benign visual variations (\eg, camera motion or changes in lighting).
A matched reference frame from the \emph{same video and similar scene} can provide a baseline, enabling the VLM to attribute meaningful, unusual differences to glitches rather than to normal gameplay variation.
RQ1 asks whether such a reference is useful when the reference itself is reliable.
Importantly, we formulate RQ1 as an \emph{oracle-reference evaluation} which provides an upper-bound condition that isolates whether a VLM can benefit from a relevant visual reference when one is available.

\paragraph{Design.}
We used \godot to evaluate RQ1 under controlled conditions.
Each test frame is evaluated under three prompting strategies:
\reref (with a relevant reference frame that was manually cherry-picked from a similar scene),
\noref (test frame only),
and \irref (with a reference frame unrelated to the evaluated gameplay video, with no meaningful connection to the game, scene, or objects).

\paragraph{Findings.}
\textbf{Oracle reference guidance improves overall detection for most VLMs, with the largest gains in recall.}
\Cref{tab:RQ1} shows that \reref increases recall and F1 for all evaluated models.
The improvement is particularly large for Qwen3-VL-8B, which gains +0.11 accuracy and +0.32 F1, driven by a +0.49 increase in recall.
Substantial improvements are also observed for much larger proprietary models such as GPT-5, which gains +0.12 accuracy and +0.25 F1, primarily due to a +0.30 increase in recall with only a -0.01 change in precision.

Gemini 3 Flash Preview is the main exception in terms of accuracy.
Although \reref increases its recall by +0.37, it also introduces substantially more false positives, resulting in a -0.04 change in accuracy.
One plausible explanation is that the model over-emphasizes discrepancies between the reference and test frames and interprets benign differences (\eg, minor viewpoint or UI changes) as evidence of a glitch.

\textbf{Irrelevant reference frames often hurt performance, confirming that reference matching is critical.}
\Cref{tab:RQ1} shows that for the smaller open models, \irref is always worse than the baseline.
For instance, Qwen3-VL-8B drops by -0.08 accuracy and -0.23 F1, driven by a -0.17 recall decrease.
Gemma-3-12B shows a similar pattern, with -0.07 accuracy and -0.34 recall, yielding a large -0.34 F1 drop.
In contrast, GPT-5 remains relatively stable from \noref to \irref, and Gemini 3 Flash Preview does not degrade.
One plausible reason is that stronger instruction-following models can down-weight the irrelevant reference when it is uninformative, whereas more reference-dependent models anchor their judgment to the irrelevant reference and get misled by scene-inconsistent comparisons.

\textbf{Oracle reference guidance improves detection across all five glitch types.}
\Cref{fig:overall_performance} shows consistent category-level benefits from \reref.
Detection improves for lighting, floating, clipping, and missing-object glitches in 5/5 models, and for texture glitches in 4/5 models; the remaining model already detects 100\% of texture glitches both with and without a reference.
Lighting glitches exhibit particularly strong and consistent gains because subtle global appearance changes can be difficult to identify from a single frame but become more salient when compared against a visual baseline.
Floating and clipping also benefit consistently because a scene-matched reference helps establish the expected geometric and physical configuration of the environment.
Improvements for texture and missing object glitches are somewhat less uniform because they depend more on whether the relevant object or material is visible in both frames and on the model's ability to perform fine-grained visual comparison under small viewpoint changes.

\begin{rqbox}
\small
\textbf{RQ1 takeaway:} A manually selected reference provides an \emph{oracle upper bound} for reference-based frame-level glitch detection. 
This oracle condition yields gains of up to +0.32 F1 and +0.12 accuracy.
Category-wise, oracle references improve lighting, floating, clipping,
and missing object in 5/5 models, and improve texture in 4/5 models.
\end{rqbox}

\subsection{RQ2: Among simple baselines, which strategy best selects a reference frame automatically?}
\label{sec:RQ2}

\paragraph{Motivation.}
RQ1 establishes that a manually selected reference can substantially improve frame-level glitch detection, providing an oracle upper bound on the potential benefit of reference guidance.
However, such manually selected references are not available in practical deployment.
The remaining challenge is therefore to identify a useful reference automatically from earlier frames in the video.
RQ2 studies this problem by establishing four simple automatic reference-selection baselines and quantifying how much of the oracle-reference benefit they can recover.

\paragraph{Design.}
We evaluate four strategies for automatically selecting a reference frame on \godot:
(1) \Last, which selects the most recent glitch-free frame;
(2) \Previous, which selects the immediately preceding frame;
(3) \Random, which samples a reference randomly from earlier frames; and
(4) \Sim, which selects the visually most similar previous frame.
We also include \noref as the baseline, and \reref as the oracle upper bound.
All strategies use our sequential prompting outputs from earlier frames to decide which frame is treated as the reference frame.

We first compare these strategies using the Qwen3-VL-8B backbone.
This choice is motivated by practical QA constraints: video analysis often requires many VLM calls per clip, so deploying a large proprietary model like GPT-5 is often prohibitively slow and costly at scale.
Qwen3VL-8B offers a more realistic performance--cost trade-off for deployment, and it exhibits a large benefit from reference guidance in~\Cref{tab:RQ1}, making it a strong backbone for studying how reference selection affects downstream performance.
After identifying the strongest baseline, we apply its selected references across all VLMs to evaluate their robustness across models.

\begin{table}[!tb]
\centering
\scriptsize
\setlength{\tabcolsep}{6pt}
\renewcommand{\arraystretch}{1.15}
\begin{tabular}{lcccc}
\toprule
\textbf{Strategy} & \textbf{Accuracy} & \textbf{F1} & \textbf{Precision} & \textbf{Recall} \\
\midrule
\reref  & 0.73 & 0.74 & 0.72 & 0.76 \\
\Last  & 0.70 & 0.69 & 0.70 & 0.69 \\
\Previous & 0.61 & 0.57 & 0.64 & 0.51 \\
\Random   & 0.67 & 0.66 & 0.68 & 0.63 \\
\Sim &0.60 &0.54 &0.64 &0.47 \\
\noref    & 0.62 & 0.42 & 0.88 & 0.28 \\
\bottomrule
\end{tabular}
\caption{Performance of reference selection strategies using Qwen3-VL-8B on \godot. \reref denotes the manually selected reference frame used in~\Cref{sec:RQ1}.}
\label{tab:RQ2}
\end{table}
\paragraph{Findings.}
\textbf{\Last performs best among the four automatic baselines, but remains below the oracle-reference condition.}
\Cref{tab:RQ2} shows that \Last outperforms \Previous, \Random, \Sim, and \noref on the Qwen3-VL-8B backbone.
One possible reason is that \Last satisfies two important conditions for an effective reference: it is likely to be clean, and it is temporally close to the test frame. Temporal proximity helps preserve scene layout, camera position, object configuration, and animation state, while the clean-frame constraint makes the comparison more meaningful for detecting abnormal visual changes.

In contrast, \Previous performs substantially worse. Although the immediately previous frame is often visually similar to the test frame, temporal adjacency alone does not guarantee that the reference is useful. The previous frame may itself contain a glitch, or may provide an ambiguous comparison. When both the reference and test frames contain the same abnormal pattern, the abnormality is no longer contrastive, making the model recognize the glitch from the test frame itself rather than from a clean-versus-glitchy comparison. This may explain why \Previous improves recall over \noref, but still performs far below \Last.

A particularly surprising result is the poor performance of \Sim, suggesting that selecting the visually most similar previous frame is insufficient as an effective reference. 
Although visual similarity can retrieve frames with similar global appearance, object layout, or scene content, it does not guarantee the reference is clean. If the most similar frame is glitchy, or contains the same artifact as the test frame, then the comparison may provide little useful contrast. Moreover, visual similarity measured by a generic image embedding may not capture the alignment needed for glitch detection, such as matching the correct animation, object state, or local motion context.

\textbf{\Last transfers across VLM families, but the oracle gap remains.}
After selecting \Last as the best automatic baseline, we re-evaluate it across multiple VLMs in \Cref{tab:RQ1}.
\Last improves F1 relative to \noref for all models and improves accuracy for 4/5 models, indicating that its benefit transfers beyond the Qwen3-VL-8B backbone used for baseline comparison.
However, the remaining gap between \Last and \reref reinforces the central benchmark challenge revealed by RQ2: automatically identifying a reference that is both clean and relevant remains an open problem.

\begin{rqbox}
\small
\textbf{RQ2 takeaway:} 
Among four simple automatic reference-selection baselines, \Last performs best and improves F1 over \noref for 5/5 VLMs and accuracy for 4/5 VLMs.
Together with the oracle upper bound established in RQ1, these results identify reliable joint reference retrieval and verification as an important open direction for future work.
\end{rqbox}

\subsection{RQ3: Does automatic reference guidance transfer to real-world gameplay data?}
\label{sec:RQ3}
\paragraph{Motivation.}
RQ1 establishes the potential benefit of reference guidance under an oracle condition.
RQ2 then evaluates four initial strategies for obtaining such references automatically on the controlled \godot benchmark, with \Last emerging as the strongest baseline.
RQ3 examines whether this benefit transfers beyond the controlled benchmark to real-world gameplay videos, where visual conditions are less constrained, scene transitions are more diverse, and suitable references may be more difficult to identify.
This setting also provides a natural stress test of reference reliability.
\Last relies on predictions from earlier frames rather than ground-truth labels: a frame predicted to be glitch-free may actually contain a glitch and can subsequently be selected as a reference.
Such errors may therefore propagate to later predictions.
We consequently evaluate both the overall frame-level benefit of automatic reference guidance and the downstream effect of correct versus incorrect references.

\paragraph{Design.} 
We evaluate the 1,000 manually annotated frames from our real-world dataset under two conditions: \Last and \noref.
To understand the reliability of automatic reference selection, we additionally inspect the ground-truth labels of the references selected by \Last after inference.
This post-hoc analysis allows us to quantify both how often automatic selection supplies an unreliable reference and how such reference errors affect downstream test-frame predictions.

\begin{table}[!tb]
\centering
\scriptsize
\setlength{\tabcolsep}{6pt}
\renewcommand{\arraystretch}{1.15}
\begin{tabular}{lcccc}
\toprule
\textbf{Strategy} & \textbf{Accuracy} & \textbf{F1} & \textbf{Precision} & \textbf{Recall} \\
\midrule

\Last & 0.60 & 0.70 & 0.56 & 0.91 \\
\noref   & 0.54 & 0.67 & 0.52 & 0.92 \\
\bottomrule
\end{tabular}
\caption{Performance of \Last and \noref on the manually labeled real-world subset of 1,000 frames.}
\label{tab:nearest_vs_noref_realworld}
\end{table}

\paragraph{Findings.}
\textbf{Automatic reference guidance remains useful on real-world data, although its effectiveness depends on reference reliability.}
As shown in~\Cref{tab:nearest_vs_noref_realworld}, on the manually labeled real-world subset, \Last achieves higher accuracy and F1 than \noref. The improvement comes mainly from higher precision, while recall remains comparable for both settings. 
These results provide initial evidence that the benefit of reference guidance is not confined to the controlled \godot environment and can transfer to gameplay videos collected independently of our synthetic generation procedure.

However, automatically selected references are imperfect. As shown in~\Cref{tab:error_analysis}, among the 1,000 evaluated frames, 551 are evaluated with a selected reference, while 449 have no reference because they are either the first frame or no prior frame has been predicted as glitch-free and is therefore available for reference selection.
Among the 551 reference-conditioned cases, 367 use a correct glitch-free reference, whereas 184 use a wrong reference whose ground-truth label is glitchy, corresponding to a wrong-reference use rate of 33.4\%. 
Hence, although \Last is effective, it introduces error propagation: frames that are incorrectly predicted as glitch-free are stored in the reference pool and used as references for later predictions.

To assess how reference reliability affects downstream prediction, we further conduct a class-conditioned analysis using true positive rate (TPR) on glitchy test frames and false positive rate (FPR) on glitch-free test frames. For glitchy test frames, wrong references reduce TPR from 0.94 to 0.82, indicating that glitchy references can make the model less sensitive to actual glitches. For glitch-free test frames, wrong references increase FPR from 0.58 to 0.84, showing that unreliable references can also cause normal frames to be misclassified as glitchy. Together, these results demonstrate that reference quality directly affects downstream prediction reliability.


\begin{table}[t]
\centering
\scriptsize
\setlength{\tabcolsep}{6pt}
\renewcommand{\arraystretch}{1.15}
\begin{tabular}{lccccc}
\toprule
\textbf{Group} & \textbf{\# Frames} & \textbf{GT=1} & \textbf{GT=0} & \textbf{TPR} & \textbf{FPR} \\
\midrule
\Last & 1,000 & 500 & 500 & 0.91 & 0.71\\
\hspace{1em} Has a reference & 551 & 303 & 248 & 0.88 & 0.62 \\
\hspace{2em} Glitch-free reference & 367 & 156 & 211 & 0.94 & 0.58 \\
\hspace{2em} Glitchy reference & 184 & 147 & 37 & 0.82 & 0.84 \\
\hspace{1em} Has no reference & 449 & 197 & 252 & 0.94 & 0.78 \\
\bottomrule
\end{tabular}
\caption{Reference reliability analysis. TPR is measured on glitchy frames, and FPR is measured on glitch-free frames.}
\label{tab:error_analysis}
\end{table}

\begin{rqbox}
\small
\textbf{RQ3 takeaway:}
Automatic reference guidance transfers beyond \godot, improving frame-level accuracy on real-world gameplay data.
However, its reliability depends on the quality of the selected references, identifying reference reliability and error propagation as key open challenges.
\end{rqbox}

\section{Conclusion}

We introduced \papertitle, a benchmark for reference-based glitch detection with VLMs. Our central motivation is that many glitches are difficult to identify from an isolated frame: a test frame may appear plausible on its own but become clearly abnormal when compared with an appropriate visual reference. To study this setting systematically, \papertitle provides a controlled synthetic dataset spanning five glitch types with manually annotated reference/test frame pairs, enabling an oracle-reference evaluation that isolates the potential benefit of reference guidance.
Across five VLMs, our experiments show that oracle references generally improve frame-level glitch detection. We further establish four initial baselines for automatic reference selection and find that \Last performs best, while still leaving a substantial gap to the oracle references. Finally, experiments on real-world gameplay data show that automatic reference guidance can improve frame-level glitch detection beyond the controlled synthetic setting, while also revealing reference reliability and error propagation as key challenges.
Together, these findings position \papertitle as a foundation for future research on reference-based game QA. Promising directions include more reliable joint reference retrieval and verification, conservative reference-pool management, and broader evaluation across diverse games, engines, glitch types, and visual styles.

{
    \small
    \bibliographystyle{ieeenat_fullname}
    \bibliography{main}
}

\end{document}